\documentclass{article}

\newcommand{\tabincell}[2]{\begin{tabular}{@{}#1@{}}#2\end{tabular}}
\usepackage{footmisc}

\usepackage[nonatbib,final]{nips_2016}

\usepackage{epsfig}
\usepackage{graphicx}
\usepackage{amsmath}
\usepackage{amssymb}
\usepackage{amsthm}
\usepackage{graphicx}
\usepackage{slashbox}
\usepackage{subfigure}
\usepackage{rotating}

\usepackage{amssymb}
\usepackage{bm}

\usepackage{epstopdf}
\usepackage{stmaryrd}
\usepackage{amsfonts}
\usepackage{array}
\usepackage{amssymb}
\usepackage{algorithmic}
\usepackage{algorithm}
\usepackage{multirow}
\usepackage{color}
\usepackage{threeparttable}
\usepackage{enumerate}
\usepackage{mathrsfs}

\usepackage{tabularx}
\usepackage{makecell, rotating}

\usepackage{setspace}
\usepackage{chngpage}
\usepackage{graphics}
\usepackage{graphicx}
\usepackage{epsfig}

\usepackage[english]{babel}
\usepackage{blindtext}
\usepackage{algorithm} 
\usepackage{algorithmic} 

\def\bD{\mathbf{D}}

\def\bU{\mathbf{U}}
\def\bV{\mathbf{V}}

\def\bY{\mathbf{Y}}

\def\bv{\mathbf{v}}
\def\bw{\mathbf{w}}
\def\bx{\mathbf{x}}

\def\b0{\mathbf{0}}
\def\b1{\mathbf{1}}

\makeatletter
\DeclareRobustCommand\onedot{\futurelet\@let@token\@onedot}
\def\@onedot{\ifx\@let@token.\else.\null\fi\xspace}

\def\etal{\emph{et al}\onedot}

\title{Self-Paced Learning: an Implicit Regularization Perspective}

\author{Yanbo Fan$^{\dagger}$ 
		\quad Ran He$^{\ast,\dagger,\ddagger}$ 
	    \quad Jian Liang$^{\ast,\dagger}$
        \quad Bao-Gang Hu$^{\dagger}$ \\
		\small{$^{\ast}$Center for Research on Intelligent Perception and Computing, CASIA}\\
		\small{$^{\dagger}$National Laboratory of Pattern Recognition, CASIA}\\
		\small{$^{\ddagger}$Center for Excellence in Brain Science and Intelligence Technology, CAS}\\
	    \texttt{\{yanbo.fan, rhe, jian.liang, hubg\}@nlpr.ia.ac.cn}
}

\begin{document}

\maketitle
%

\begin{abstract}

	Self-paced learning (SPL) mimics the cognitive mechanism of humans and animals that gradually learns from easy to hard samples. One key issue in SPL is to obtain better weighting strategy that is determined by minimizer function. Existing methods usually pursue this by artificially designing the explicit form of SPL regularizer. In this paper, we focus on the minimizer function, and study a group of new regularizer, named self-paced implicit regularizer that is deduced from robust loss function. 
	Based on the convex conjugacy theory, the minimizer function for self-paced implicit regularizer can be directly learned from the latent loss function, while the analytic form of the regularizer can be even known. 
	A general framework (named SPL-IR) for SPL is developed accordingly. 
	We demonstrate that the learning procedure of SPL-IR is associated with latent robust loss functions, thus can provide some theoretical inspirations for its working mechanism.
	We further analyze the relation between SPL-IR and half-quadratic optimization. 
	Finally, we implement SPL-IR to both supervised and unsupervised tasks, and  
	experimental results corroborate our ideas and demonstrate the correctness and effectiveness of implicit regularizers.
	%

\end{abstract}

\section{Introduction}
%
Inspired by the learning process and cognitive mechanism of humans and animals,  
Bengio \etal propose a new learning strategy called \emph{curriculum learning} (CL) in \cite{bengio2009curriculum}, which gradually includes more and more hard samples into training process.
A curriculum can be seen as a sequence of training criteria.
For example, in the training of a shape recognition system, images that exhibit less variability such as squares and circles are considered first, followed by hard shapes like ellipses.
%
The curriculum in CL is usually determined by some certain priors, and thus is problem specific and lacks generalizations.
To alleviate this, Kumar \etal propose a new learning strategy named self-paced learning (SPL) that incorporates the curriculum updating in the process of model optimization \cite{kumar2010self}. 
General SPL model consists of a problem specific weighted loss term on all samples and a SPL regularizer on sample weights. Alternative search strategy (ASS) is generally used for optimization.
%
By gradually increasing the penalty of the SPL regularizer during the optimization, more samples are included into training from easy to hard by a self-paced manner.
%
%
Due to its ability of avoiding bad local minima and improving the generalization performance, many works have been developed based on SPL 
\cite{li2016multi,liang2016,jiang2015self,zhang2015self,supancic2013self,lee2011learning}.
%
%
%

One key issue in SPL is to obtain better weighting strategy that is determined by the minimizer functions, and existing methods usually pursue this by artificially designing the explicit form of SPL regularizers \cite{xu2015multi,zhao2015self,jiang2014easy,jiang2014self}.
Some examples are listed in the appendix.
Specifically, a definition of self-paced regularizer is given in \cite{jiang2014easy}.
Though shown to be effective in many applications experimentally, the underlying working mechanism of SPL is still unclear and is heavily desired for its future development. 
One attempt in this aspect is \cite{meng2015objective}, they show that the ASS method used for SPL accords with the \emph{majorization minimization} \cite{vaida2005parameter} algorithm implemented on a latent SPL objective, and deduce the latent objective of hard, linear and mixture regulraizers.

%
%

Considering the crucial role of minimizer function in SPL, we focus on it and study a group of new regularizer (named self-paced implicit regularizer) for SPL based on the convex conjugacy theory. 
Comparing with existing SPL regularizers, the self-paced implicit regularizer is deduced from robust loss function and its analytic form can be even unknown. 
Its properties and corresponding minimizer function can be learned from the latent loss function directly.
Besides, the proposed self-paced implicit regularizer is independent of the learning objective and thus leads to a general framework (named SPL-IR) for SPL. SPL-IR can be optimized via ASS algorithm.
More importantly, we demonstrate that the learning procedure of SPL-IR is indeed associated with latent robust loss functions, thus may provide some theoretical inspirations for its working mechanism (e.g. its robustness to outliers and heavy noise). 
\textcolor{black}{
We further analyze the relations between SPL-IR and half-quadratic (HQ) optimization and provide a group of self-paced implicit regularizer accordingly. Such relations can be beneficial to both SPL and HQ optimization.
}
Finally, we implement SPL-IR to three classical tasks (i.e. matrix factorization, clustering and classification). Experimental results corroborate our ideas and demonstrate the correctness and effectiveness of SPL-IR.

Our work has three main contributions:
(1) We propose self-paced implicit regularizer for SPL, and develop a general implicit regularization framework (named SPL-IR) based on it. The self-paced implicit regularizers not only enrich the family of regularizers for SPL but also can provide some inspirations on the working mechanism of SPL.
%
%
(2) \textcolor{black}{We analyze the connections between SPL-IR and HQ optimization}, and provide a group of robust loss function induced self-paced implicit regularizers for SPL-IR accordingly. 
%
(3) Experimental results on both supervised and unsupervised tasks corroborate our ideas and demonstrate the correctness and effectiveness of SPL-IR.
\section{Preliminaries}


\subsection{Self-Paced Learning via Explicit Regularizers}
%
Given training dataset \( \bD = \{ (\bx_i,y_i )\}_{i=1}^{n} \) with $n$ samples, where $\bx_{i} \in R^d$ is the $i$-th sample, $y_i$ is the optional information according to the learning objective (e.g.  $y_i$ can be the label of $\bx_i$ in classification model). Let $f(. \ ,\bw)$ denote the learned model and $\bw$ be the model parameter. 
$L(y_{i},f(\bx_{i},\bw))$ is the loss function of $i$-th sample. 
%
%

\textcolor{black}{
Mimicking the cognitive mechanism of humans and animals, SPL aims to optimize the model from easy to hard samples gradually. 
}
The objective of SPL is to jointly optimize model parameter $\bw$ and latent sample weights $\bv = [v_1,v_2,\dots,v_n]$ via the following minimization problem:
\begin{equation}
\label{self_paced_function}
\min_{\bw,\bv} \ \mathbb{E}(\bw,\bv; \lambda) = \sum_{i=1}^{n} v_{i} L(y_{i},f(\bx_{i},\bw )) + g(\lambda,v_i),
\end{equation}
where $g(\lambda,v)$ is called self-paced regularizer and $\lambda$ is a penalty parameter that controls the learning pace. 
ASS algorithm is generally used for (\ref{self_paced_function}), which alternatively optimizes $\bw$ and $\bv$ while keeping the other fixed.
Specifically, given sample weights $\bv$, the minimization over $\bw$ is a weighted loss minimization problem that is independent of regularizer $g(\lambda,v)$; given model parameter $\bw$, the optimal weight of $i$-th sample is determined by
\begin{equation}
\min_{v_i} \quad v_i L(y_{i},f(\bx_{i},\bw )) + g(\lambda, v_i).
\end{equation} 
Since $\ell_i = L(y_{i},f(\bx_{i},\bw )$ is constant once $\bw$ is given, the optimal value of $v_i$ is uniquely determined by the corresponding minimizer function $\sigma(\lambda, \ell_i)$ that satisfies
\begin{equation}
\sigma(\lambda, \ell_i) \ell_i + g(\lambda, \sigma(\lambda, \ell_i)) \leq v_i \ell_i + g(\lambda, v_i), \forall v_i \in [0,1].
\end{equation}
%
For example, if $g(\lambda, v_i) = -\lambda v_i$ \cite{kumar2010self}, the optimal $v_i^*$ is calculated by
\begin{eqnarray}
	v_i^* = \sigma(\lambda, \ell_i ) = \left\{
	\begin{array}{ll}
	1, \quad if \quad \ell_i \le \lambda \\
	0, \quad otherwise
	\end{array}
	\right.
	\label{spl_minimization}
\end{eqnarray}
By gradually increasing the value of $\lambda$, more and more hard samples are included into the training process.
Many efforts have been put into the learning of minimizer functions \cite{xu2015multi,zhao2015self,jiang2014easy,jiang2014self,supancic2013self}, and we name them as SPL with explicit regularizers as they usually require the explicit form of regularizer $g(\lambda,v)$
. 
$\sigma(\lambda,\ell)$ is then derived from the form of $g(\lambda,v)$.

\begin{table*}
\footnotesize
\centering
\caption{   
\small{
Loss function $\phi(\lambda,t)$ and the corresponding minimizer function $\sigma(\lambda,t)$, $\lambda$ is a hyper-parameter. }
}
\label{implicit_spl_regularizer}
\setlength\tabcolsep{1pt}
\begin{tabular}{|c|c|c|c|c|}

%
\hline
	& Huber	& Cauchy & L1-L2 & Welsch\\

\hline
Loss function $\phi(\lambda,t)$  
& 	
$\left\{ {\begin{array}{*{20}{c}}
	{t^2/2,}&{|t| \leq \lambda }\\
	{\lambda|t| - \frac{\lambda^2}{2},}&{|t| > \lambda}
	\end{array}} 
\right.$

&
$\lambda^2 \log(1+(t/\lambda)^2)$ 

&
$\sqrt{\lambda + t^2} - 1$ 

&
$\lambda^2(1 -\exp(-\frac{t^2}{\lambda^2}))$\\

\hline
Minimizer function $\sigma(\lambda,t)$ 

&
$\left\{ {\begin{array}{*{20}{c}}
	{1}&{|t| \leq \lambda }\\
	{\lambda/|t|,}&{|t| > \lambda}
	\end{array}} 
\right.$

&
$2 / (1+(t/\lambda)^2)$ 

&
$1 / \sqrt{\lambda + t^2}$ 

&
$2 \exp(-\frac{t^2}{\lambda^2})$ \\

\hline
\end{tabular}
\end{table*}

\subsection{Half-Quadratic Optimization}
Half-quadratic optimization \cite{nikolova2005analysis,geman1995nonlinear,geman1992constrained} is a commonly used optimization method that based on the convex conjugacy theory. It tries to solve a nonlinear objective function via optimizing a series of half-quadratic reformulation problems iteratively \cite{he2014robust,he2014half,he2011maximum,he2010two,yuan2009robust}. 
%

Given a differentiable function $\phi(t): R \to R$, if $\phi(t)$ further satisfies the conditions of the multiplicative form of HQ optimization in \cite{nikolova2007equivalence}, the following equation holds for any fixed $t$, 
\begin{equation}
\phi(t) = \inf_{p \in R_+} \left\{  \frac{1}{2} pt^2 + \psi(p) \right\}, 
\label{hq_funtion}
\end{equation}
where $\psi(p)$ is the dual potential function of $\phi(t)$ and $R_+ = \{t| t \geq 0\}$. $\psi(p)$ is convex and reads
\begin{equation}
	\psi(p) = \sup_{t \in R_+} \left\{ -\frac{1}{2} pt^2 + \phi(t) \right\}, 
	\label{dpf_funtion}
\end{equation}
More analysis about $\phi(t)$ and $\psi(p)$ refers to \cite{nikolova2005analysis}. The optimal $p^*$ that minimize (\ref{hq_funtion}) is uniquely determined by the corresponding minimizer function $\delta(t)$
, 
which is derived from convex conjugacy and is only relative to function $\phi(t)$. For each $t$, $\delta(t)$ is such that
\begin{equation}
\frac{1}{2} \delta(t) t^2 + \psi(\delta(t)) \leq \frac{1}{2} pt^2 + \psi(p), \ \forall p \in R_+.
\label{mini_func}
\end{equation}
The optimization of $\phi(t)$ can be done via iteratively minimizing $t$ and $p$ in (\ref{hq_funtion}). 
One only needs to focus on $\phi(t)$ and its corresponding minimizer function $\delta(t)$ in HQ optimization, 
and the analytical form of the dual potential function $\psi(p)$ can be even unknown.  

%

\section{The Proposed Method}


%
In this section, we first give the definition of the proposed self-paced implicit regularizer and derive its minimizer function based on convex conjugacy. Then we develop a general self-paced learning framework, named SPL-IR, based on implicit regularization. Finally, we analyze the relations between SPL-IR and HQ optimization.

\subsection{Self-Paced Implicit Regularizer}

Based on our above analysis of SPL, we define the self-paced implicit regularizer as follows,
%
%
\begin{spacing}{1.2}
\end{spacing}
\textbf{Definition 1. }\emph{\textbf{Self-Paced Implicit Regularizer.}}
\emph{A self-paced implicit regularizer $\psi(\lambda,v)$ is defined as the dual potential function of a robust loss function $\phi(\lambda,t)$, 
and satisfies}
\begin{spacing}{1.1}
	\emph{1. \(\phi(\lambda,t) = \min_{v \geq 0} \ vt + \psi(\lambda,v)\); } 
	
	\emph{2. $\sigma(\lambda,t)$ is the minimizer function of $\phi(\lambda,t)$ that satisfies
	$\sigma(\lambda, t) t + \psi(\lambda, \sigma(\lambda, t)) \leq v t + \psi(\lambda, v)$, $\forall \ v \in R_+$;
	}
	
	\emph{3. \(\sigma(\lambda,t) \) is non-negative and up-bounded, $\forall \ t \in R_+$;}
	
	\emph{4. \(\sigma(\lambda,t) \) is monotonically decreasing w.r.t. $t$, $\forall \ t \in R_+$;}
	
	\emph{5. \(\sigma(\lambda,t) \) is monotonous w.r.t. $\lambda \in R_+$;} \\
	\emph{where $\lambda$ is a hyper-parameter and it is the same in  \(\phi(\lambda,t)\), \(\psi(\lambda,v)\) and $\sigma(\lambda,t)$. 
	$\lambda$ is considered to be fixed in the first four conditions.}
\end{spacing}

\begin{spacing}{1.2}
\end{spacing}
\textbf{Proposition  1}
\emph{
For any fixed $\lambda$, if $\phi(\lambda,t)$ in Definition 1 further satisfies the conditions referred in \cite{nikolova2007equivalence}, its minimizer function $\sigma(\lambda,t)$ is uniquely determined by $\phi(\lambda,t)$ and the analytic form of the dual potential function \(\psi(\lambda,v)\) can be even unknown during the optimization.
}
\begin{spacing}{1.2}
\end{spacing}
The proof of Proposition 1 is given in the appendix. 
According to Definition 1, the self-paced implicit regularizer is derived from robust loss function. Its properties can be learned from both $\psi(\lambda,v)$ and the latent loss function $\phi(\lambda,t)$. 
The corresponding minimizer function $\sigma(\lambda,t)$ can be learned from $\phi(\lambda,t)$ directly. 
During the optimization, the optimal $v^*$ is determined by $\sigma(\lambda,t)$ and the analytic form of $\psi(\lambda,v)$ can be even unknown,
hence $\psi(\lambda,v)$ is named self-paced implicit regularizer.
Besides, the last three conditions in Definition 1 are required for SPL regimes. 
Specifically, let $t$ denote the sample loss, condition 4 indicates that the model is likely to select easy samples (with smaller losses) in favor of hard samples (with larger losses) for a fixed $\lambda$, and condition 5 makes sure that we can incorporate more and more samples through turning parameter $\lambda$.

Besides, Jiang \etal have given a definition of self-paced regularizer and derived necessary conditions of the regularizer and the corresponding minimizer function for SPL in \cite{jiang2014easy}. However, it is still nontrivial to design self-paced regularizers or analyze their properties accordingly.  
The self-paced implicit regularizer $\psi(\lambda,v)$ defined here is derived from robust loss function $\phi(\lambda,t)$.
By establishing the relations between $\phi(\lambda,t)$ and $\psi(\lambda,v)$, we can analyze their working mechanisms as well as develop new SPL regularizers based on the development of robust loss functions. 
Moreover, the properties of $\psi(\lambda,v)$ and its corresponding minimizer function $\sigma(\lambda,t)$ can be learned from $\phi(\lambda,t)$.



\subsection{Self-Paced Learning via Implicit Regularizers}
We can develop an implicit regularization framework for SPL based on the proposed self-paced implicit regularizer. By substituting the regularization term $g(\lambda,v)$ in (\ref{self_paced_function}) with a self-paced implicit regularizer $\psi(\lambda,v)$ given in Definition 1, we obtain the following SPL-IR problem,
\begin{equation}
\min_{\bw,\bv} \ \mathbb{E}(\bw,\bv; \lambda) = \sum_{i=1}^{n} v_{i} L(y_{i},f(\bx_{i},\bw )) + \psi(\lambda,v_i).
\label{spir_spl}
\end{equation}
It can be solved via ASS algorithm, which alternatively optimizes $\bw$ and $\bv$ while keeping the other fixed. However, different from existing SPL regularizers, the analytic form of $\psi(\lambda,v)$ in (\ref{spir_spl}) can be unknown and the optimal $\bv^*$ is determined by the corresponding minimizer function given in Definition 1. 
The optimization procedure of (\ref{spir_spl}) is described in Algorithm \ref{alg_for_spir}. 
Model (\ref{spir_spl}) is called an implicit regularization framework since it does not require the explicit form of $\psi(\lambda,v)$.
The benefit of implicit regularization has been analyzed in \cite{mahoney2012approximate,orecchia2011implementing}.

\begin{figure}[!tb]
	\setlength{\abovecaptionskip}{0pt}
	\setlength{\belowcaptionskip}{-15pt}
	\centering
	\subfigure{\label{subfig:loss}
		\includegraphics[height=2in]{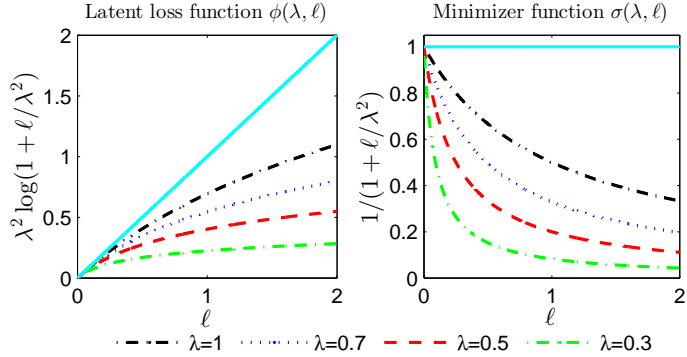}	
		}
	\caption{\small{Example of latent loss function and its corresponding minimizer function in Definition 1. The x-axis refers to original loss $\ell$. The solid lines are given for comparison, it is $y = x$ in left figure, and $y = 1$ in right one.}}
	\label{fig:cauchy}
\end{figure}

An insightful phenomenon is that the learning procedure of SPL-IR is actually associated with certain latent loss functions. 
For example, for a certain implicit regularizer and its corresponding minimizer function $v^*_i = \sigma(\lambda,\ell_i) = 1 / (1 + \ell_i / \lambda^2)$ in Algorithm \ref{alg_for_spir} (where $\ell_i = L(y_{i},f(\bx_{i},\bw^* ))$), 
one is actually minimizing a latent robust function $\sum_{i=1}^{n} \lambda^2 \log(1 + \ell_i /\lambda^2)$  during each round.
%
Figure \ref{fig:cauchy} gives a graphical illustration.
The latent loss function $\phi(\lambda,\ell)$ can be considered to carry out a meaningful transformation on original loss $\ell$. When $\ell$ is larger than a certain threshold, $\phi(\lambda,\ell)$ becomes a constant and its corresponding minimizer function $\sigma(\lambda,\ell)$ becomes zero, hence the related sample is not considered for optimization.
Through this, it can suppress the influence of hard samples (refer to larger $\ell$) while retaining that of easy samples (refer to smaller $\ell$).
This may also provide some inspirations on the robustness of SPL-IR to outliers and heavy noise as they can usually cause larger losses.
More specifically, 
starting with a small $\lambda$ (e.g. 0.3), only a small part of samples with very small losses will be involved (they are considered to contain reliable information). As $\lambda$ increases, the suppressing effect of $\phi(\lambda,\ell)$ on larger losses becomes weaker and their corresponding weights increase, consequently more and more hard samples with larger losses (may also contain more knowledge) are involved into training process.
While gradually incorporating these knowledge, the model becomes stronger and stronger.
The learning procedure of some existing regularizers like hard and linear \cite{meng2015objective} can also be explained under the framework of SPL-IR.


%
%

SPL-IR in (\ref{spir_spl}) is considered as a general SPL framework from two aspects: firstly,  $\psi(\lambda,v)$ represents a spectrum of self-paced implicit regularizer that is developed based on robust loss function and convex conjugacy theory; secondly, $\psi(\lambda,v)$ is independent of specific model objective $L(y_{i},f(\bx_{i},\bw ))$ and thus can be used in various applications. 
Besides, standard ASS strategy is used for both SPL with explicit regularizer (model (\ref{self_paced_function})) and SPL-IR (model (\ref{spir_spl})). It includes a weighted loss minimization step and a weight updating step at each iteration, and the time overhead is mainly in the former step. Hence for a specific loss function $L(y_{i},f(\bx_{i},\bw ))$ and a fixed number of iteration, the time complexities of SPL with explicit regularizer and SPL-IR is in the same order of magnitude.
%



%

\begin{figure}[!tb]
	\setlength{\abovecaptionskip}{0pt}
	\setlength{\belowcaptionskip}{-15pt}
	\centering
	\subfigure[Toy Example]{\label{subfig:loss}
		\includegraphics[height=1.5in]{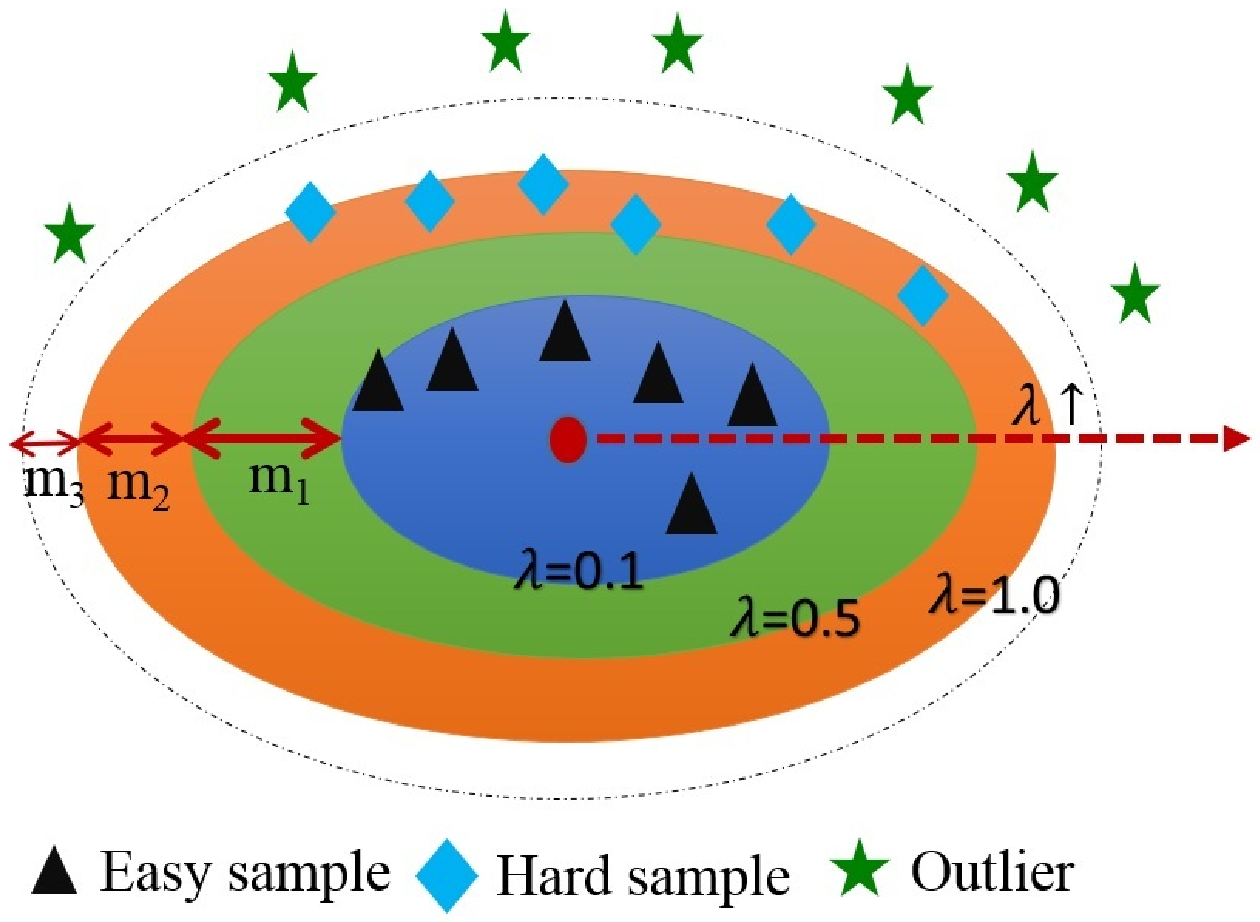}
	}
	\subfigure[HQ and SPL-IR]{\label{hq_spir_welsch}
		\includegraphics[height=1.5in]{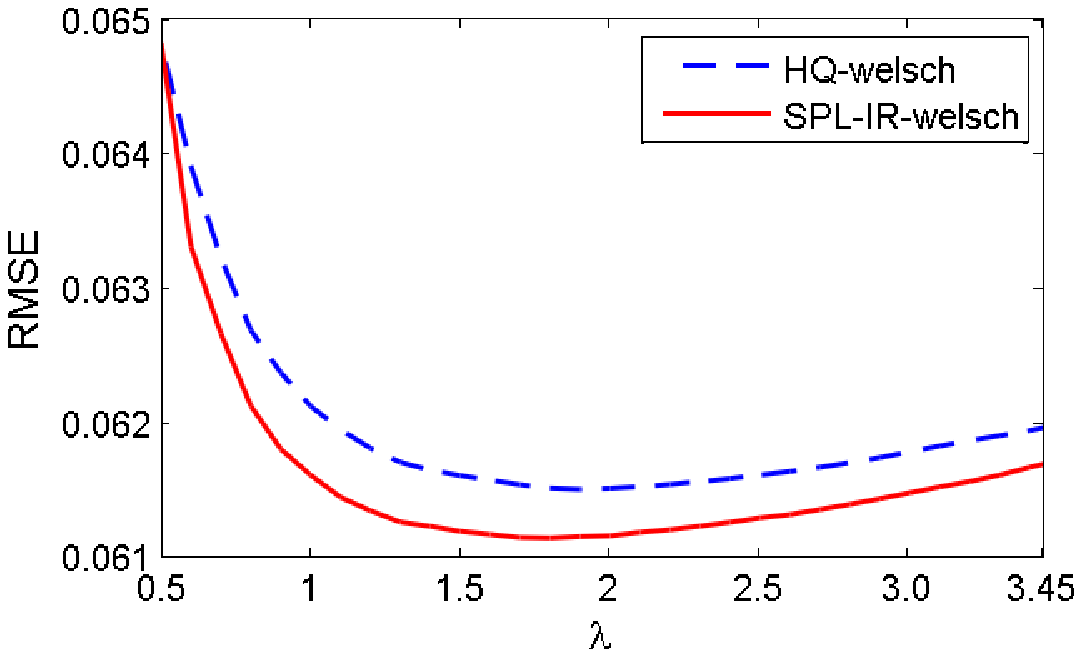}
		}
	
	\caption{	
	\small{	
	In (a), training samples are roughly divided into three types: easy samples $\blacktriangle$, hard samples $\blacklozenge$ and outliers $\bigstar$. 
	$\lambda$ is usually fixed in HQ methods (e.g. $\lambda = 0.5$), hence some samples may be discarded incorrectly. 
	In contrast, SPL-IR can gradually incorporate more samples from easy to hard (i.e. $\lambda$ grows iteratively). 	
	(b) demonstrates the performances of HQ and SPL-IR methods on a synthetic matrix factorization dataset, Welsch minimizer function is adopted for both methods. For HQ-welsch, standard HQ algorithm \cite{nikolova2005analysis} is implemented with each $\lambda$ independently. More details refer to Section 3.3 and 4.1.
	}}
	\label{fig:huber}
\end{figure}


%

\subsection{SPL-IR and Half-Quadratic Optimization}

We can develop new self-paced implicit regularizers based on the development of robust loss functions. Specifically, 
we analyze the relations between SPL-IR and HQ optmization and provide several self-paced implicit regularizers accordingly. For better demonstration, we first give an equivalent quadratic form definition of self-paced implicit regularizer
,
\begin{spacing}{1.2}
\end{spacing}
\textbf{Definition 2 (Quadratic Form).} 
\emph{\textbf{Self-Paced Implicit Regularizer.}}
\emph{A self-paced implicit regularizer $\psi(\lambda,v)$ is defined as the dual potential function of a robust loss function $\phi(\lambda,t)$, 
and satisfies}
\begin{spacing}{1.1}
	\emph{1. \(\phi(\lambda,t) = \min_{v \geq 0} \ \frac{1}{2}\ vt^2 + \psi(\lambda,v)\); } 
	
	\emph{2. $\sigma(\lambda,t)$ is the minimizer function of $\phi(\lambda,t)$ and satisfies
	$\frac{1}{2} \sigma(\lambda, t) t^2 + \psi(\lambda, \sigma(\lambda, t)) \leq 
	\frac{1}{2} v t^2 + \psi(\lambda, v)$, $\forall \ v \in R_+.$
	}
	
	\emph{3. \(\sigma(\lambda,t) \) is non-negative and up-bounded, $\forall \ t \in R_+$;}
	
	\emph{4. \(\sigma(\lambda,t) \) is monotonically decreasing w.r.t. $t$, $\forall \ t \in R_+$;}
	
	\emph{5. \(\sigma(\lambda,t) \) is monotonous w.r.t. $\lambda \in R_+$;} \\
	\emph{where $\lambda$ is a hyper-parameter and it is the same in  \(\phi(\lambda,t)\), \(\psi(\lambda,v)\) and $\sigma(\lambda,t)$. 
		$\lambda$ is considered to be fixed in the first four conditions.}
\end{spacing}
\begin{spacing}{1.2}
\end{spacing}

\begin{algorithm}
	\caption{: Self-Paced Learning via Implicit Regularizers}
	\label{alg_for_spir}
	\begin{algorithmic}[1]
		\REQUIRE Input dataset $\bD = \{\bx_i, y_i\}_{i=1}^{n}$, step size $\mu > 1$.
		\ENSURE  Model parameter $\bw$.
		
		\STATE Initialize sample weights $\bv^*$ and parameter $\lambda$;
		
		\REPEAT
		
		\STATE Update $(\bw^*,\bv^*) = \arg\min_{\bw,\bv} \ \mathbb{E}(\bw,\bv; \lambda) $ by using ASS	algorithms, $\bv$ is iteratively optimized by the corresponding minimizer function $\sigma$;

		\STATE Monotone increase (or decrease) $\lambda$ by step-size $\mu$;
		
		\UNTIL {convergence}.
		\RETURN $\bw*$ 
	\end{algorithmic}
\end{algorithm}

%
%
%
%
%
%

\begin{table*}[!htbh]
	\footnotesize
	\centering
	\caption{
	\small{Numerical results of $L_1$-norm MF problem with $L_2$-norm regularization. The best results are highlighted in bold.}}
	\label{result_mf_l1_l2}
	\setlength\tabcolsep{2.8pt}
	\begin{tabular}{c|c c c c c c c}
		\hline
		Method & \tabincell{c} {PRMF} & \tabincell{c}{SPL-hard}  & 
		\tabincell{c}{SPL-mixture} & 
		\tabincell{c}{SPL-IR-huber} & \tabincell{c}{SPL-IR-L1-L2} &
		\tabincell{c}{SPL-IR-cauchy} & \tabincell{c}{SPL-IR-welsch} \\

		\hline
		
		RMSE & 0.1528 & 0.0949 & 0.0625 & 0.0627 & 0.0650 & 0.0620 & \textbf{0.0596}\\
		\hline
		
		MAE  & 0.0994 & 0.0672 & 0.0475 & 0.0476 & 0.0493 & 0.0472 & \textbf{0.0455}\\
		
		\hline
	\end{tabular}
\end{table*}

%
%
%
%
%
%
%
The equivalency of Definition 1 and Definition 2 is shown in the appendix.
Seen from Definition 2, there is a close relationship between self-paced implicit regularizer and the dual potential function defined in HQ reformulation (\ref{hq_funtion}). 
Apparently, the dual potential function in (\ref{hq_funtion}) and the minimizer function in (\ref{mini_func}) satisfy the first two conditions in Definition 2,
\textcolor{black}{
and self-paced implicit regularizer imposes further constraints on the minimizer function $\sigma(\lambda,t)$ for the regimes of SPL.
} 
Many loss functions and their corresponding minimizer functions in multiplicative form of HQ have been developed (some of them are tabulated in Table \ref{implicit_spl_regularizer}). It is easy to verify that the functions in Table \ref{implicit_spl_regularizer} satisfy all the conditions in Definition 2, hence they can be adjusted for self-paced implicit regularizers.
The loss functions in Table \ref{implicit_spl_regularizer} are well defined and have proven to be effective in many areas \cite{he2014half}.
Meanwhile, though self-paced implicit regularizer can be developed from HQ optimization, their optimization procedures are quite different. In HQ, one mainly focuses on the minimization of loss function $\phi(\lambda,t)$ and hyper-parameter $\lambda$ is predetermined and fixed during the optimization. While aiming to gradually optimize from easy to hard samples, SPL-IR uses the right-hand side $vt^2/2 + \psi(\lambda,v)$ to model problems and one key concern is the weighting strategy that determined by the minimizer function $\sigma(\lambda,t)$.
Besides, in order to gradually increase samples, $\lambda$ is updated stage by stage in SPL-IR.

Figure \ref{fig:huber} gives an intuitive interpretation. 
If we set $t_i = \sqrt{L(y_{i},f(\bx_{i},\bw^* ))}$ and use the minimizer function of Welsch given in Table \ref{implicit_spl_regularizer} for weight updating in Algorithm \ref{alg_for_spir}, model (\ref{spir_spl}) can be considered to sequential optimize a group of Welsch loss functions with monotonically increasing $\lambda$. 
Hence SPL-IR is able to gradually optimize from easy to hard samples while incorporating the good properties of robust Welsch functions. 
On the other hand, for HQ optimization, $\lambda$ is predefined and fixed during the whole optimization. 
Hence its performance may be largely influenced by the selection of $\lambda$.
For example, when $\lambda$ is somehow small (e.g. $\lambda < 1$ in Figure~\ref{hq_spir_welsch}), some hard samples will be simply considered as outliers and discarded. 
From the comparisons in Figure~\ref{hq_spir_welsch}, we can find that SPL-IR can always outperform HQ for every $\lambda$.

\section{Experiments}

To illustrate the correctness and effectiveness of the developed SPL-IR model, we apply it to three classical tasks: matrix factorization, clustering and classification. Experimental results  demonstrate that the proposed self-paced implicit regularizers outperform baseline algorithms and achieve comparable or even better performance comparing to the artificially designed SPL regularizers.


There are two hyper-parameter $(\lambda,\mu)$ that need to be tuned in Algorithm \ref{alg_for_spir}.
We follow a standard setting in SPL \cite{kumar2010self} for all our experiments. That is, $\lambda$ is initialized to obtain about half samples, then it is iteratively updated to involve more and more samples gradually. 
The practical updating direction depends on the specific minimizer function. For functions given in Table \ref{implicit_spl_regularizer}, $\lambda_{T+1} = \lambda_T / \mu$ for L1-L2 while $\lambda_{T+1} = \lambda_T * \mu$ for Huber, Cauchy and Welsch, where $\mu > 1$ is a step factor and $T$ is an iteration number. $\mu$ is empirically set to 1.05 in our experiments.
Similar settings are adjusted for the competing SPL regularizers,
including SPL-hard \cite{kumar2010self} and SPL-mixture \cite{zhao2015self}.
%
\begin{figure}[!tb]
	\setlength{\abovecaptionskip}{0pt}
	\setlength{\belowcaptionskip}{-10pt}
	\centering
	\subfigure{
		\includegraphics[height=1.5in]{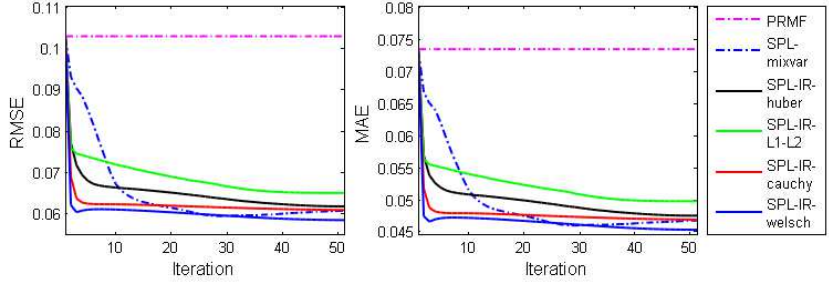}
	}

	\caption{\small{Tendency curves of RMSE and MAE w.r.t. the iterations.}}
	\label{fig:mf_result}
\end{figure}

\begin{table*}[!tb]
	\footnotesize
	\centering
	\caption{
		\small{Clustering performance on the Handwritten Digit dataset. The best results are highlighted in bold.}
	}
	\label{result_mvc}
	\begin{tabular}{c| c c c c c}

		\hline
		Method & ACC & NMI & AR & F-score & Purity\\
		\hline
		\hline
		
		FOU & 0.612(0.066) &	 0.628(0.029) &	 0.484(0.049) &	 0.539(0.043) &	 0.645(0.051) \\
		\hline
		
		FAC & 0.588(0.044) &	 0.597(0.017) &	 0.453(0.031) &	 0.512(0.027) &	 0.631(0.032) \\
		\hline
		
		KAR & 0.734(0.062) &	 0.730(0.030) &	 0.634(0.055) &	 0.672(0.049) &	 0.767(0.048) \\
		\hline
		
		MOR & 0.415(0.014) &	 0.500(0.003) &	 0.295(0.004) &	 0.374(0.003) &	 0.475(0.004) \\
		\hline
		
		PIX & 0.677(0.059) &	 0.701(0.031) &	 0.585(0.050) &	 0.629(0.045) &	 0.711(0.047) \\
		\hline
		
		ZER & 0.524(0.033) &	 0.504(0.016) &	 0.369(0.024) &	 0.434(0.021) &	 0.551(0.022) \\
		\hline
		\hline
			
		Con-MC & 0.775(0.078) &	 0.773(0.037) &	 0.690(0.066) &	 0.722(0.058) &	 0.802(0.059) \\
		\hline
		
		SPL-hard & 0.821(0.059) &	 0.758(0.029) &	 0.709(0.050) &	 0.739(0.044) &	 0.834(0.045) \\
		\hline
		
		SPL-mixture & 0.845(0.068)	 & 0.812(0.030)	& 0.763(0.057)	& 0.787(0.051)	& 0.861(0.050) \\
		\hline
		
		MSPL & 0.840(0.070) &	 0.806(0.035) &	 0.751(0.064) &	 0.776(0.057) &	 0.854(0.054)  \\
		\hline
		
		SPL-IR-huber & 0.843(0.070)	& 0.810(0.035)	& 0.756(0.064)	& 0.781(0.057)	& 0.858(0.053) \\
		\hline

		SPL-IR-L1-L2 & 0.835(0.068)	& 0.801(0.034)	& 0.743(0.061)	& 0.769(0.054)	& 0.849(0.052)	 \\
		\hline
		
		SPL-IR-cauchy & 0.845(0.071) & 0.814(0.035) &	 0.762(0.064) &	 0.786(0.057) &	 0.861(0.053) \\
		\hline
		
		SPL-IR-welsch & \textbf{0.862(0.071)}	& \textbf{0.833(0.035)}	& \textbf{0.790(0.064)}	& \textbf{0.812(0.057)}	& \textbf{0.878(0.053)}	 \\
		
		\hline
	\end{tabular}
\end{table*}

\subsection{Matrix Factorization}
Matrix factorization (MF) is one of the fundamental problems in machine learning and data mining. It aims to factorize an $m \times n$ data matrix $\bY$ into two smaller factors $\bU \in R^{m \times r}$ and $\bV \in R^{n \times r}$, where $r \ll min(m,n)$, such that $\bU \bV^T$ is possibly close to $\bY$.
MF has been successfully implemented in many applications, such as collaborative filtering \cite{SalMnih08}.

Here we consider the MF problem on synthetic dataset. Specifically, the data used here is generated as follows: two matrices $\bU$ and $\bV$, both of which are of size $100 \times 4$, are first randomly generated with each entry drawn from the Gaussian distribution $\mathcal{N}(0,1)$, leading to a ground truth rank-$4$ matrix $\bY_0 = \bU \bV^T$. Then we randomly choose $40\%$ of the entries and treat them as missing data. Another 20\% of the entries are randomly selected and added to uniform noise on $[-20,20]$, and the rest are perturbed with Gaussian noise drawn from $\mathcal{N}(0,0.1^2)$.
Similar to \cite{zhao2015self}, we consider $L_1$-norm MF problem with $L_2$-norm regularization,
and the baseline algorithm is PRMF \cite{wang2012probabilistic}.
We modify it with different SPL regularizers for comparison.
%
%
%
Two commonly used metrics are adopted here: (1) \emph{root mean square error} (RMSE): $\frac{1}{\sqrt{mn}} ||\bY_0 - \hat{\bU} \hat{\bV}^T||_F$, and (2) \emph{mean absolute error} (MAE):$\frac{1}{mn} ||\bY_0 - \hat{\bU} \hat{\bV}^T||_1$, where $\hat{\bU}$ and $\hat{\bV}$ denote the outputs of MF algorithms. 
All the algorithms are implemented with 50 realizations and their mean values are reported.
Table \ref{result_mf_l1_l2} tabulates their numerical results. All SPL-IR algorithms obtain performance improvements over baseline algorithm PRMF, which shows the benefits of SPL regimes. 
Comparing among different SPL regularizers, the results of proposed self-paced implicit regularizers are comparable to or even better than that of mixture and hard schemes, especially for SPL-IR with welsch regularizer. 
These demonstrate the correctness and effectiveness of the proposed self-paced implicit regularizer.
%
Figure \ref{fig:mf_result} further plots the tendency curves of RMSE and MAE with different self-paced implicit regularizers and mixture regularizer for better understanding, the results of PRMF are also reported as a baseline.
The performances of all implicit regularizers improve rapidly for the first few iterations as more and more easy samples are likely to be involved in these phases. With the increasing of the iterations, the improvements become steady as some hard instances or outliers are included.

\subsection{Multi-view Clustering}
Multi-view clustering aims to group data with multiple views into their underlying classes \cite{xu2013survey}. Most existing multi-view clustering algorithms fit a non-convex model and may be stuck in bad local minima. To alleviate this, Xu \etal propose a multi-view self-paced learning algorithm (MSPL) that considers the learnability of both samples and views and achieves promising results in \cite{xu2015multi}.
Here we simply modified their MSPL model with different SPL regularizers for comparison.
%
%
%
The UCI Handwritten Digit dataset \footnote{https://archive.ics.uci.edu/ml/datasets \label{uci}} is used in this experiment. It consists of 2,000 handwritten digits classified into ten categories (0-9).
Each instance is represented in terms of the following six kinds of features (or views): Fourier coefficients of the character shapes (FOU), profile correlations (FAC), Karhunen-Love coefficients (KAR), pixel averages in 2 x 3 windows (PIX), Zernike moments (ZER), and morphological features (MOR). Here we make use of all the six views for all the comparing algorithms.
The baseline algorithms are standard k-means on each single view's representation and Con-MC (the features are concatenated on all views firstly, and then standard k-means is applied). 

Five commonly used metrics are adopted to measure the clustering performances: clustering accuracy (ACC), normalized mutual information (NMI), F-score, Purity, and adjusted rand index (AR) \cite{hubert1985comparing}. Higher value indicates better performance for all the metrics.
All algorithms are implemented 20 times and both mean values and standard derivations are reported. Table \ref{result_mvc} tabulates their numerical results.
It can be seen that all the multi-view algorithms obtain significant improvements over single-view ones, which demonstrates the benefits of integrating information from different views. 
More importantly, comparing to Con-MC, the SPL-IR algorithms can further improve the performance by gradually optimizing from easy to hard samples and avoiding bad local minima. 
The proposed self-paced implicit regularizers are comparable to or even better than the compared SPL regularizers. 

\begin{table} [!tb]
	\footnotesize
	\caption{\small{Statistical Information of Databases.}}
	\centering
	\label{statistical_information_database}
	\setlength\tabcolsep{8pt}
\begin{tabular}{c| c c c }
		\hline
		\textbf{Dataset} & \textbf{\#.Category} & \textbf{\#.Instance} & \textbf{\#.Feature} \\
		
		\hline
		Breast & 2 & 569 & 30 \\
				
		\hline
		Spambase & 2 & 4601 & 57 \\
		
		\hline
		Svmguide1 & 2 & 7089 & 4 \\
				
		\hline
	\end{tabular}
\end{table}

\begin{table*}[!tb]
	\footnotesize
	\centering
	\caption{\small{Classification accuracy (\%).}}
	\label{result_classification}
	\setlength\tabcolsep{3.5pt}
	\begin{tabular}{c| c c c c c c c}
		\hline
		\multicolumn{8}{c}{Without Label Noise} \\
		
		\hline
		Method & \tabincell{c} {LR }  & \tabincell{c}{SPL-\\hard} & \tabincell{c}{SPL-\\mixture} & 
		\tabincell{c}{SPL-IR-\\huber} & \tabincell{c}{SPL-IR-\\L1-L2} &
		\tabincell{c}{SPL-IR-\\cauchy} & \tabincell{c}{SPL-IR-\\welsch} \\
		\hline
		
		Breast  & 97.36(2.22) & 97.54(2.22) &	 98.25(1.65) 	& \textbf{98.77(1.19)} 	& 97.90(1.79) 	& 98.42(1.54) 	& 98.25(1.65) 	\\ 
		Spambase  & 92.35(1.47) & 92.63(1.08) & 92.83(1.44) 	& 93.05(1.25) 	& 93.00(1.36) 	& 93.09(1.41) 	& \textbf{93.13(1.34)} 	\\ 
		Svmguide1 & 95.39(0.95) & 95.39(0.95) &  95.51(1.04) 	& 95.57(0.95) 	& 95.57(1.10) 	& 95.65(1.01) 	& \textbf{95.68(0.90)} 	\\ 
		\hline
		
%
%
%

		\hline
		\multicolumn{8}{c}{With 20\% Random Label Noise} \\
		
		\hline
		Method & \tabincell{c} {LR } & \tabincell{c}{SPL-\\hard} & \tabincell{c}{SPL-\\mixture} & 
		\tabincell{c}{SPL-IR-\\huber} & \tabincell{c}{SPL-IR-\\L1-L2} &
		\tabincell{c}{SPL-IR-\\cauchy} & \tabincell{c}{SPL-IR-\\welsch} \\
		\hline
		Breast  & 92.08(2.96) 	& 96.13(2.15) & 96.66(2.12) 	& 96.84(2.33) 	& 94.72(2.89) 	& 97.54(1.90) 	& \textbf{97.89(1.63)} 	\\ 
		Spambase  & 89.28(1.66)  & 89.81(1.61) & 90.76(1.82) 	&  90.92(1.65)	& 90.09(1.65) 	& 90.85(1.55) 	& \textbf{91.37(1.37)} 	\\ 
		Svmguide1 & 91.52(0.65)  & 92.72(1.12) & 93.81(0.79) 	& 93.54(0.75) 	& 92.83(0.71) 	& 93.88(1.05) 	& \textbf{94.37(0.90)} 	\\ 
		
		\hline
		
	\end{tabular}
\end{table*}
\subsection{Classification}
The proposed self-paced implicit regularizers can be flexible implemented to supervised tasks.
Here we conduct a binary classification task.
Specifically, we utilize the L2-regularized Logistic Regression (LR) model as our baseline, and incorporate it with different SPL regularizers for comparison.
Liblinear \cite{fan2008liblinear} is used as the solver of LR.
Three real-world databases are considered: Breast\footref{uci}, Spambase\footref{uci} and Svmguide1 \cite{chang2011libsvm}. 
Their statistical information is summarized in Table \ref{statistical_information_database}.
For each dataset, we consider it without additional noise and with 20\% random label noise, respectively.
The 20\% random label noise means we randomly select 20\% samples from training data and reversal their labels (change positive to negative, and vice-versa). 
We use 10-fold cross validation for all the databases, and report both their mean values and their standard derivations.

Classification accuracy is used for performance measure.
Table \ref{result_classification} reports their numerical results. 
For both situations, SPL-IR algorithms can get performance improvements over original LR method to some extent.
Moreover, when adding random label noise, the performance of original LR degenerates a lot, while the SPL algorithms can still obtain relatively high performance, especially for SPL-IR with welsch regularizer.
This corroborates our analysis about the robustness of SPL-IR to outliers and heavy noise.



\section{Conclusions}
In this paper, we study a group of new regularizer, named self-paced implicit regularizer for SPL based on the convex conjugate theory. 
The self-paced implicit regularizer is derived from robust loss function and its analytic form can be even unknown.
Its properties and the corresponding minimizer function can be learned from the latent loss function directly.
We then develop a general SPL framework (SPL-IR) based on it.
We further demonstrate that the learning procedure of SPL-IR is actually associated with certain latent robust loss functions,
thus may provide some theoretical inspirations on the working mechanisms of SPL-IR (such as the robustness to outliers or heavy noise). 
We later analyze the relations between SPL-IR and HQ optimization and develop a group of self-paced implicit regularizer accordingly.
Experimental results on both supervised and unsupervised tasks demonstrate the correctness and effectiveness the proposed self-paced implicit regularizer.


\section{Appendix}

\subsection{Proof of Proposition 1}
\textbf{Proof. }
The proof sketch is similar to that in \cite{nikolova2007equivalence}. For ease of representation, we omit $\lambda$ and use $\phi(t)$, $\psi(v)$ and $\sigma(t)$ for short. Some fundamental assumptions about $\phi(t)$ are: 
\textbf{H1:} $\phi: R_{+} \rightarrow R$ is increasing with $\phi \not\equiv 0$ and $\phi(0) =0$;
\textbf{H2:} $\phi(t)$ is $C^1$ and concave;
\textbf{H3:} $\lim_{t \rightarrow \infty} \phi(t) / t = 0$.

Put $\theta(t) = -\phi(t)$, then $\theta$ is convex by {H2}. Its convex conjugate is 
$\theta^*(v) = \sup_{t \geq 0} \ \{vt - \theta(t)\}.$
By the Fenchel-Moreau theorem \cite{rockafellar2015convex}, the convex conjugate of $\theta^*$ is $\theta$, that is 
$
\theta(t)  = (\theta^*)^*(t) = \sup_{v \leq 0} \ \{vt - \theta^*(v)\}  = -\inf_{v \geq 0} \ \{vt + \theta^*(-v)\}.
$
Thus we have 
\begin{equation}
\psi(v) = \theta^*(-v) = \sup_{t \geq 0} \ \{-vt - \theta(t)\}
 = \sup_{t \geq 0} \ \{-vt + \phi(t)\}.
\label{dul_1}
\end{equation}
\begin{equation}
\phi(t) = -\theta(t) = \inf_{v \geq 0} \ \{vt + \theta^*(-v)\}
 = \inf_{v \geq 0} \ \{vt + \psi(v)\}.
 \label{ori_1}
\end{equation}
Then the problem becomes how to achieve the supremum in (\ref{dul_1}) jointly with the infimum in (\ref{ori_1}).
For any $\hat{v} > 0$, define $ f_{\hat{v}}: R_{+} \rightarrow R$ by $f_{\hat{v}}(t) = \hat{v}t + \theta(t)$, 
then we have $\psi(\hat{v}) = - \inf_{t \geq 0} f_{\hat{v}}(t)$ from (\ref{dul_1}). 
According to {H1-H3}, $f_{\hat{v}}$ is convex with $f_{\hat{v}}(0) =0$ and $\lim_{t \rightarrow + \infty} f_{\hat{v}}(t) = + \infty$.
Thus $f_{\hat{v}}$ can reach its unique minimum at a $\hat{t} \geq 0$, and $\psi(\hat{v}) = -\hat{v} \hat{t} + \phi({\hat{t}})$ from (\ref{dul_1}). Hence equivalently the infimum in (\ref{ori_1}) is reached at $\hat{v}$ as 
$\phi({\hat{t}}) = \hat{v} \hat{t} + \psi(\hat{v})$.
\textcolor{black}{
Then we have $\hat{v} = \sigma(t) = -\theta'(t) = \phi'(t).$
}
Thus the optimal $v$ is uniquely determined by the minimizer function $\sigma(t)$ that is derived from $\phi(t)$. The analytic form of the dual potential function $\psi(v)$ could be unknown during the optimization. 
The proof is then completed.

%

\subsection{Definition 1 and Definition 2}
To show the equivalency of Definition 1 and Definition 2 in the main body, we first give the following proposition about Definition 2.

\begin{spacing}{1.2}
\end{spacing}
\textbf{Proposition 2}
\emph{
For any fixed $\lambda$, if $\phi(\lambda,t)$ in Definition 2 further satisfies the conditions referred in \cite{nikolova2007equivalence}, its minimizer function $\sigma(\lambda,t)$ is uniquely determined by $\phi(\lambda,t)$ and the analytic form of \(\psi(\lambda,v)\) can be even unknown during the optimization.
}
\begin{spacing}{1.4}
\end{spacing}
Proof. The proof sketch is similar to that in \cite{nikolova2007equivalence}. For ease of representation, we omit $\lambda$ and use $\phi(t)$, $\psi(v)$ and $\sigma(t)$ for short. Some fundamental assumptions about $\phi(t)$ are: 
\textbf{H1:} $\phi: R_{+} \rightarrow R$ is increasing with $\phi \not\equiv 0$ and $\phi(0) =0$;
\textbf{H2:} $t \rightarrow \phi(\sqrt{t})$ is concave;
\textbf{H3:} $\phi(t)$ is $C^1$;
\textbf{H4:} $\lim_{t \rightarrow \infty} \phi(t) / t^2 = 0$.

Put $\theta(t) = -\phi(\sqrt{t})$, then $\theta$ is convex by {H2}. Its convex conjugate is 
$\theta^*(v) = \sup_{t \geq 0} \ \{vt - \theta(t)\}.$
By the Fenchel-Moreau theorem \cite{rockafellar2015convex}, the convex conjugate of $\theta^*$ is $\theta$, that is 
$
\theta(t)  = (\theta^*)^*(t) = \sup_{v \leq 0} \ \{vt - \theta^*(v)\}  = -\inf_{v \geq 0} \ \{vt + \theta^*(-v)\}.
$
Define $\psi(v) = \theta^*(-\frac{1}{2}v)$, we have
\begin{equation}
\psi(v) = \sup_{t \geq 0} \ \{-\frac{1}{2}vt - \theta(t)\}
 = \sup_{t \geq 0} \ \{-\frac{1}{2}vt^2 + \phi(t)\}.
\label{dul}
\end{equation}
\begin{equation}
\phi(t) = -\theta(t^2) = \inf_{v \geq 0} \ \{vt^2 + \theta^*(-v)\}
 = \inf_{v \geq 0} \ \{\frac{1}{2}vt^2 + \psi(v)\}.
 \label{ori}
\end{equation}
Then the problem becomes how to achieve the supremum in (\ref{dul}) jointly with the infimum in (\ref{ori}).
For any $\hat{v} > 0$, define $ f_{\hat{v}}: R_{+} \rightarrow R$ by $f_{\hat{v}}(t) = \frac{1}{2}\hat{v}t + \theta(t)$, 
then we have $\psi(\hat{v}) = - \inf_{t \geq 0} f_{\hat{v}}(t)$ from (\ref{dul}). 
According to {H1-H4}, $f_{\hat{v}}$ is convex with $f_{\hat{v}}(0) = 0$ and $\lim_{t \rightarrow + \infty} f_{\hat{v}}(t) = + \infty$.
Thus $f_{\hat{v}}$ can reach its unique minimum at a $\hat{t} \geq 0$, and $\psi(\hat{v}) = -\frac{1}{2}\hat{v} \hat{t}^2 + \phi({\hat{t}})$ from (\ref{dul}). Hence equivalently the infimum in (\ref{ori}) is reached at $\hat{v}$ as 
$\phi({\hat{t}}) = \frac{1}{2}\hat{v} \hat{t}^2 + \psi(\hat{v})$.
\textcolor{black}{
Then we have $\hat{v} = \sigma(t) = -2\theta'(t^2) = \phi'(t)/t.$
}
Thus the optimal $v$ is uniquely determined by the minimizer function $\sigma(t)$ that is only related to $\phi(t)$. The analytic form of the dual potential function $\psi(v)$ could be unknown during the optimization. 
The proof is then completed.

\begin{spacing}{1.2}
\end{spacing}
Denote $\ell_i = L(y_{i},f(\bx_{i},\bw ))$ and rewrite model (8) in the main body as 
\begin{equation}
\min_{\bw,\bv} \ \mathbb{E}(\bw,\bv; \lambda) = \sum_{i=1}^{n} v_{i} (\sqrt{\ell_i})^2 + \psi(\lambda,v_i).
\label{spir_spl_2}
\end{equation}
If we adopt $\psi(\lambda,v_i)$ with an implicit regularizer given in Definition 2 and use $v_i^* = \frac{1}{2} \sigma(\lambda,\sqrt{\ell_i})$, where $\sigma(\lambda,\sqrt{\ell_i})$ is the minimizer function in Definition 2, model (\ref{spir_spl_2}) is optimizing a latent loss function $ \sum_{i=1}^{n}\phi(\lambda, \sqrt{\ell_i})$ equivalently. 

\begin{spacing}{2}
\end{spacing}

Now we demonstrate the equivalency of Definition 1 and Definition 2 in the main body. 
For easy of representation, we omit $\lambda$, and use $\{\phi_1(t), \psi_1(v), \sigma_1(t)\}$ and $\{\phi_2(t), \psi_2(v), \sigma_2(t)\}$ to refer to the functions in Definition 1 and Definition 2, respectively. 
Considering a simplified model
\begin{equation}
\min_{\bw,v} \ vL(y,f(\bx,\bw )) + \psi(v).
\label{simply_model}
\end{equation}
Denote $\ell = L(y,f(\bx,\bw ))$. We show that for a same implicit regularizer $\psi(v) = \psi_1(v) = \psi_2(v)$, the  optimal $v^*$ and the latent loss function of model (\ref{simply_model}) derived from Definition 1 and Definition 2 are the same.
Specifically, let $\psi_1(v) = \psi_2(v) = \sup_{t \geq 0} \ \{-vt + \phi_1(t)\}$ (where $\phi_1(t)$ satisfies conditions H1-H3 of Proposition 1 in the main body), it is easy to verify that its corresponding latent loss function is $\phi_1(\ell)$ and optimal $v^* = \sigma_1(\ell) = \phi_1'(\ell)$ according to Definition 1 and Proposition 1.
Meanwhile, we have $\psi_2(v) = \sup_{t \geq 0} \ \{-vt + \phi_1(t)\} = \sup_{t \geq 0} \ \{-vt^2 + \phi_2(t)\}$, where $\phi_2(t) = \phi_1(t^2)$. Then 
model (\ref{simply_model}) can be considered to optimize a latent loss function $\phi_2(\sqrt{\ell}) = \phi_1(\ell)$ and the optimal $v^* = \frac{1}{2} \sigma_2(\sqrt{\ell}) = \phi_1'(\ell)$ according to Definition 2 and Proposition 2.
Thus we show the equivalency of Definition 1 and Definition 2.

\subsection{Self-Paced Regularizer}
%
Similar definitions of self-paced regularizer (or self-paced function) have been proposed in \cite{jiang2015self,zhao2015self,jiang2014easy}. The definition in \cite{zhao2015self} is shown below.

\textbf{Definition 3} 
\emph{
\textbf{(Self-Paced Regularizer)}
\cite{zhao2015self}: Suppose that $v$ is a weight variable, $\ell$ is the loss, and $\lambda$ is the learning pace parameter. $g(\lambda,v)$ is called self-paced rgularizer, if
}
\begin{spacing}{1.2}
\end{spacing}
\begin{spacing}{1.1} 
	\emph{1. \(g(\lambda,v)\) is convex with respect to $v \in [0,1]$;} 
	
	\emph{2. $v^*(\lambda, \ell)$ is monotonically decreasing w.r.t. $\ell$, and it holds that $\lim_{\ell \to 0} v^*(\lambda, \ell)$} = 1,
	\emph{$\lim_{\ell \to \infty} v^*(\lambda, \ell)$} = 0 \emph{;} 
	
	\emph{3. $v^*(\lambda, \ell)$ is monotonically increasing w.r.t. ${\lambda}$, and it holds that $\lim_{\lambda \to 0} v^*(\lambda, \ell)$} = 0, \emph{$\lim_{\lambda \to \infty} v^*(\lambda, \ell) \leq $} 1 \emph{;} 
		
	\emph{where $v^*(\lambda, \ell) = \arg\min_{v \in [0,1]} v\ell + g(\lambda,v)$.}
\end{spacing}
\begin{spacing}{1.2}
\end{spacing}

\begin{table*}[!htb]
\footnotesize
\centering
\caption{\small{Recently proposed self-paced regularizers $g(\lambda,v)$ and their corresponding $v^*(\lambda, \ell)$}}
\label{existed_spl_regularizer}
\setlength\tabcolsep{1pt}
\begin{tabular}{|l|l|l|}

%
\hline
	& $g(\lambda,v)$	& $v^*(\lambda, \ell)$ \\

\hline
Kumar \etal \cite{kumar2010self}
& 	

$-\lambda \sum_{i=1}^{n} v_i,\ \lambda>0$

&
$\left\{ {\begin{array}{*{20}{c}}
	1, &  \ell_i < \lambda \\
	0, &  otherwise
	\end{array}} 
\right.$
\\

\hline
Jiang \etal \cite{jiang2014easy,jiang2015self}

&
$\frac{1}{2} \lambda \sum_{i=1}^{n} (v_i^2 - 2v_i), \ \lambda > 0 $

&
$\left\{ {\begin{array}{*{20}{c}}
	1 - \frac{1}{\lambda}\ell_i, &  \ell_i < \lambda \\
	0, &  otherwise
	\end{array}} 
\right.$
\\

\hline
Jiang \etal \cite{jiang2014easy,jiang2015self}

&
$\begin{aligned}
&\sum_{i=1}^{n}( \zeta v_i - \frac{\zeta^{v_i}}{\log \zeta}), \\
&\zeta = 1 - \lambda, 0 < \lambda < 1 
\end{aligned}$

&
$\left\{ {\begin{array}{*{20}{c}}
	\frac{1}{\log \zeta} \log(\ell_i + \zeta), &  \ell_i < \lambda \\
	0, &  otherwise
	\end{array}} 
\right.$
\\

\hline
{Jiang \etal \cite{jiang2014easy,jiang2015self} }

&

$\begin{aligned}
&-\zeta \sum_{i=1}^{n}\log(v_i + \frac{1}{\lambda_1}\zeta), \\
&\zeta = \frac{\lambda_1 \lambda_2}{\lambda_1 - \lambda_2}, \lambda_1 > \lambda_2 >0
\end{aligned}
$

&
$\left\{ {\begin{array}{*{20}{c}}
	1, &  \ell_i \leq \lambda_2 \\
	\frac{(\lambda_1 - \ell_i)\zeta}{\ell_i \lambda_1}, & \lambda_2 < \ell_i < \lambda_1 \\
	0, &  \ell_i \geq \lambda_1
	\end{array}} 
\right.$
\\

\hline
Jiang \etal \cite{jiang2014self}

&
$-\lambda\sum_{i=1}^{n} v_i - \gamma||\bv||_{2,1}, \ \lambda > 0, \ \gamma > 0$

&
$\left\{ {\begin{array}{*{20}{c}}
	1, &  \ell_i \leq \lambda + \gamma \frac{1}{\sqrt{i} - \sqrt{i-1}} \\
	0, &  otherwise
	\end{array}} 
\right.$
\\

\hline
Xu \etal \cite{xu2015multi}

&
$\begin{aligned}
&\sum_{i=1}^{n} \ln(1 + e^{-\lambda} - v_i)^{(1 + e^{-\lambda} - v_i)} \\
&+ \ln(v_i)^{v_i} - \lambda v_i, \ \lambda > 0
\end{aligned}$

&
$\frac{1 + e^{-\lambda}}{1 + e^{\ell_i - \lambda}}$
\\

\hline
Zhao \etal \cite{zhao2015self}

&
$\sum_{i=1}^{n} \frac{\lambda \gamma^2}{\lambda v_i + \gamma}, \ \lambda > 0, \ \gamma > 0$

&
$\left\{ {\begin{array}{*{20}{c}}
	1, &  \ell_i \leq (\frac{\lambda \gamma}{\lambda + \gamma})^2 \\
	0, &  \ell_i \geq \lambda^2 \\
	\gamma (\frac{1}{\sqrt{\ell_i}} - \frac{1}{\lambda}), & otherwise 
	\end{array}} 
\right.$
\\

\hline
Zhang \etal \cite{zhang2015self}

&
$\begin{aligned}
&-\lambda \sum_{k=1}^{K} \sum_{i=1}^{n_k} v_i^k -\gamma \sum_{k=1}^{K} \sqrt{\sum_{i=1}^{n_k} v_i^k},\\
&\lambda > 0, \ \gamma > 0
\end{aligned}$

&
$\left\{ {\begin{array}{*{20}{c}}
	1, &  \ell_i^k < \lambda + \frac{\gamma}{2\sqrt{i}} \\
	\frac{((\frac{\gamma}{2(\ell_i^k-\lambda)})^2 - (i-1)}{m}, &  otherwise
	\end{array}} 
\right.$
\\
\hline
\end{tabular}
\end{table*}

Table \ref{existed_spl_regularizer} tabulates some examples of self-paced regularizers $g(\lambda,v)$ and their corresponding $v^*(\lambda, \ell)$. 
We modify their original expressions for better comparison.
It is still nontrivial to design self-paced regularizers or analyze their properties according to Definition 3. 
Besides, though shown to be effective in many applications experimentally, the underlying working mechanism of SPL is still unclear.

One attempt about the underlying working mechanism of SPL is \cite{meng2015objective}. Starting from SPL regularizers and their minimizer functions, they show that the ASS method used for SPL accords with the \emph{majorization minimization} \cite{vaida2005parameter} algorithm implemented on a latent SPL objective, and deduced the latent objective of hard, linear and mixture regulraizers.
In contrast, we start from a latent loss function $\phi(\lambda,\ell)$ directly and propose self-paced implicit regularizer based on the convex conjugacy theory. 
We establish the relations between robust loss function $\phi(\lambda,\ell)$, self-paced implicit regularizer $\psi(\lambda,v)$ and minimizer function $\sigma(\lambda,\ell)$. 
According to Definition 1, $\psi(\lambda,v)$ and $\sigma(\lambda,\ell)$ are derived from latent loss function $\phi(\lambda,\ell)$, thus we can analyze their properties based on the development of $\phi(\lambda,\ell)$ (many loss functions have be widely studied in related areas).
We further demonstrate that for SPL with the proposed implicit regularizer, its learning procedure actually associates with certain latent robust loss functions.
Thus we can provide some inspirations for the working mechanism of SPL (e.g. its robustness to outliers and heavy noise).
Moreover, by establishing the relations between $\phi(\lambda,\ell)$ and $\psi(\lambda,v)$, we can develop new SPL regularizers based on the development of robust loss functions. Specifically, we analyze the relations between self-paced implicit regularizer and HQ optimization.
Many robust loss functions and their minimizer functions have been developed and widely used in HQ optimization, and they can be adjusted for self-paced implicit regularizers (some examples are given in Table 1 in main body).
%
%

{
\small
\bibliographystyle{ieee}
\bibliography{egbib}
}

\end{document}